\documentclass[conference]{IEEEtran}
\IEEEoverridecommandlockouts
\usepackage{cite}
\usepackage{amsmath,amssymb,amsfonts}
\usepackage{algorithmic}
\usepackage{graphicx}
\usepackage{textcomp}
\usepackage{xcolor}
\usepackage{booktabs}
\usepackage[caption=false,font=footnotesize]{subfig}
\def\BibTeX{{\rm B\kern-.05em{\sc i\kern-.025em b}\kern-.08em
    T\kern-.1667em\lower.7ex\hbox{E}\kern-.125emX}}
\usepackage{capt-of}
\usepackage{caption}

\begin{document}

\title{ReGATE: Recurrent Gated Visual Fusion for Reactive Humanoid--Object Interaction
}

\author{%
  \IEEEauthorblockN{Hao Zhu and Jorge Peña-Queralta}
  \IEEEauthorblockA{%
    Centre for Artificial Intelligence, Zürich University of Applied Sciences,\\
    Winterthur, Switzerland\\
    \texttt{\{zhuh,penq\}@zhaw.ch}%
  }%
}

\twocolumn[{%
    \renewcommand{\twocolumn}[1][]{#1}%
    \maketitle

    \vspace{-0.5em}

    \begin{center}
        \includegraphics[
            width=\textwidth,
            keepaspectratio
        ]{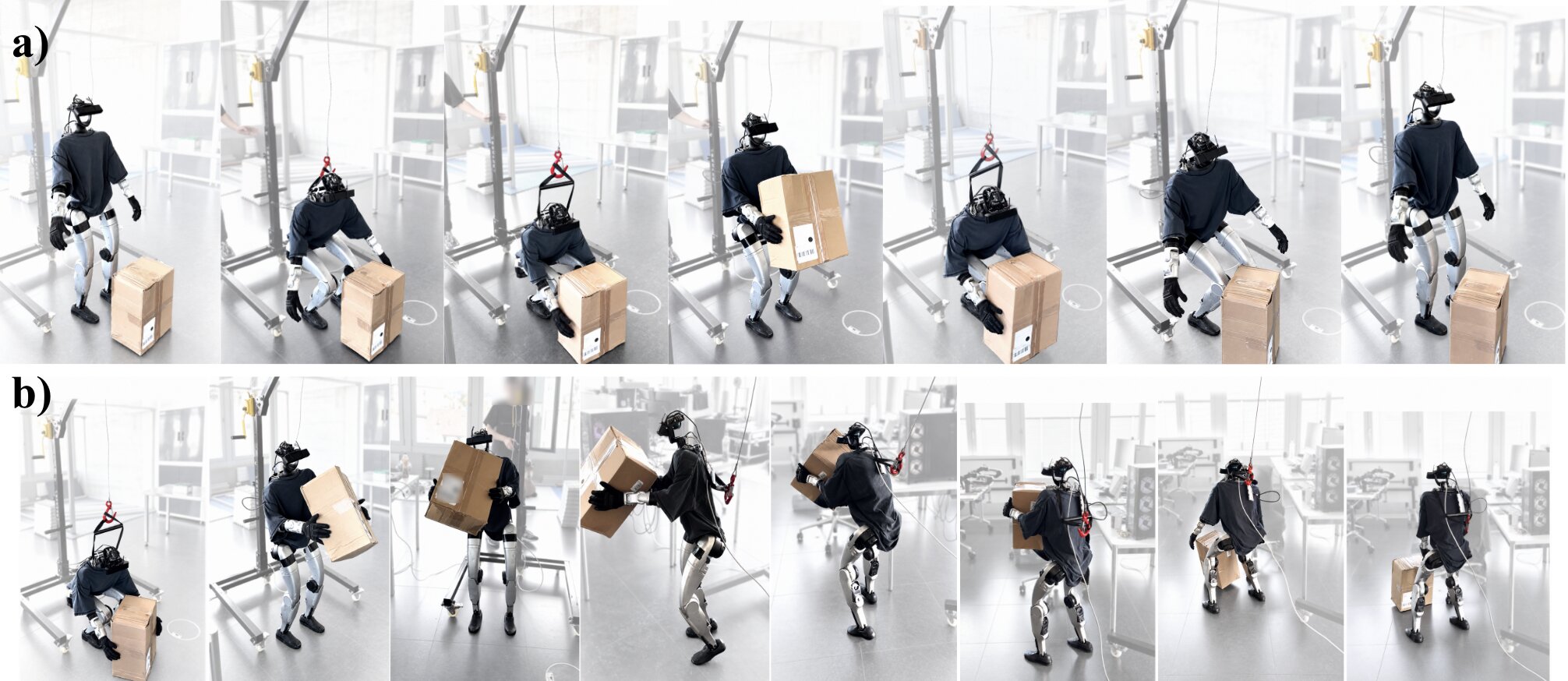}

        \vspace{-0.5em}

        \captionof{figure}{ReGATE enables command-conditioned whole-body object
        interaction on a Unitree G1 using onboard proprioception and egocentric
        depth. Frames progress from left to right.
        (a) \emph{Contact-command response}: activating the contact command
        initiates grasping and lifting, while deactivating it prompts
        the robot to place and release the box.
        (b) \emph{Commanded loco-manipulation}: after commanded grasping, the robot follows
        planar velocity commands to transport the box while maintaining balance
        and physical contact.}
        \label{fig:real_world_command}
    \end{center}

    \vspace{0.5em}
}]

\begin{abstract}

Humanoid loco-manipulation requires coordinating object acquisition, transport, and release from onboard sensing while responding to user commands and incomplete visual observations. We present ReGATE, a whole-body controller that combines egocentric depth, proprioception, and online locomotion and contact commands. Built on a teacher–student framework, ReGATE integrates interaction-state estimation through a recurrent adapter that predicts object geometry and physical contact without an external object-pose estimator. Head-wise and global gates regulate visual updates to a proprioceptive pathway. A visual-grounding and corruption curriculum trains the adapter, while phase-aware initialization increases coverage of brief contact transitions. In simulation, ReGATE achieves 85.2\% success across 330 static-object acquisition trials, compared with 48.5\% for VisualMimic under the same protocol, and 85.9\% across 440 moving-object acquisition trials. Deployment on a Unitree G1 demonstrates commanded grasp–transport–release sequences and acquisition of laterally displaced boxes. A qualitative hardware experiment further demonstrates successful box acquisition when the camera is occluded after an initial observation and before the contact command. To the best of our knowledge, this is the first work to extend recurrent gated fusion of onboard depth and proprioception from perceptive locomotion to humanoid loco-manipulation under interrupted visual sensing.

\end{abstract}


\section{Introduction}



Humanoid robots offer the potential to perform object interaction in human-centered environments \cite{wang2025physhsi,sonic2026}. Their embodiment enables coordinated whole-body behaviors that integrate locomotion, grasping, and object transport \cite{visualmimic2025,prohoi2026}.  A deployable controller must coordinate these stages from onboard sensing and support user-specified locomotion and contact commands. Learning-based methods have enabled increasingly robust humanoid--object interaction through reinforcement learning and motion imitation
\cite{boxloco2023,sugar2026,Li2026haic,sonic2026,hdmi2025}.
However, many existing systems either track predefined interaction motions~\cite{sugar2026, Li2026haic} or rely on object states obtained from external motion-capture systems~\cite{hdmi2025}, limiting their deployment outside instrumented environments. Recent methods address this limitation using perceptive closed-loop control. Some estimate object states through explicit perception pipelines~\cite{ultra2026,prohoi2026,wang2025physhsi}, while others distill privileged interaction policies into visuomotor students operating from onboard observations~\cite{vaic2026, visualmimic2025}. Despite this progress, executing a complete interaction remains challenging when the controller must perceive the object for closed-loop interaction and adapt the interaction online to user commands.

Three challenges are particularly important for robust HOI. First, obtaining object states through external perception packages introduces additional engineering dependencies and may propagate estimation errors to the controller. An end-to-end policy should instead infer the task-relevant object state directly from onboard depth and proprioceptive history. Second, the reliability of depth observations changes throughout an interaction. Occlusion by the object or the robot body, missing measurements, latency, and frozen frames can make visual evidence unreliable, requiring the controller to preserve a stable proprioceptive interaction belief while selectively incorporating useful visual information. Third, the controller must respond to user-defined locomotion and contact commands rather than rigidly replaying a reference trajectory. This requires coordinated adaptation of whole-body motion and contact behavior across different stages of object manipulation.

To tackle these challenges, we present \textbf{ReGATE}, a whole-body controller for humanoid--object interaction. ReGATE operates from onboard proprioception and egocentric depth, together with a user command specifying planar velocity, yaw rate, and desired contact state, allowing the user to adjust robot movement and contact timing during deployment. To improve coverage of complete interaction sequences, we introduce phase-aware random state initialization, which partitions demonstrations into approach, pre-contact, interaction, pre-release, and release phases. Combined with lightweight perception--action rewards, this strategy emphasizes brief transition states and encourages object acquisition and recovery after unexpected contact loss.

At the core of ReGATE is a recurrent gated fusion adapter that estimates task-relevant object state directly from depth observations and proprioceptive history. The adapter augments each depth frame with image coordinates and encodes it into spatial tokens using a convolutional encoder and recurrent visual memory. Cross-attention retrieves visual evidence conditioned on the current proprioceptive state and user command. Head-wise and global gates regulate the contribution of this fused evidence, while an always-active proprioceptive pathway maintains the interaction representation when depth becomes unreliable. The fused representation is decoded under auxiliary supervision into object pose and physical contact. These predictions are then temporally aggregated to reconstruct the interaction latent learned by the privileged teacher. A visual curriculum first grounds the depth pathway under clean observations and subsequently introduces occlusion and sensing corruption. We evaluate ReGATE on a box pickup and carrying HOI task under online command variations, contact transitions, and visual degradation. Across the tested box-picking scenarios, ReGATE localizes the object from onboard depth and completes commanded grasp--transport--release sequences under both nominal and degraded visual observations.

In summary, the contributions of this work are:
    
    

\begin{itemize}
    \item \textbf{Reactive whole-body control:} A deployable controller for humanoid loco-manipulation that operates from onboard depth and proprioception and responds to online locomotion and contact commands.

    \item \textbf{Integrated interaction-state estimation:} A recurrent adapter that predicts object geometry and physical contact from onboard observations, providing the control policy with an interaction representation without an external object-pose estimator or an object template supplied at deployment.

    \item \textbf{Gated visual fusion:} A hierarchical gating mechanism that regulates depth updates to a proprioceptive pathway, trained with a visual-grounding and corruption curriculum and demonstrated on physical object acquisition after the camera is occluded.
\end{itemize}

Together, these contributions extend recurrent gated perceptive control to humanoid loco-manipulation, including object acquisition when visual access is completely interrupted before the contact command.

\section{Related Work}
\label{sec:related_work}

\subsection{Learning Humanoid--Object Interaction}

Recent advances in  general motion tracking (GMT) have established a foundation for acquiring dynamic humanoid behaviors from motion data~\cite{peng2018deepmimic,Peng2021amp}. BeyondMimic and SONIC extend this paradigm toward versatile and scalable whole-body motion tracking~\cite{liao2025beyondmimic,sonic2026}. Recent work adapts these motion priors to contact-rich humanoid--object interaction (HOI). Box Loco-Manipulation learns unified locomotion and box transport~\cite{boxloco2023}; HAIC models coupled robot--object interaction as dynamic world model for interactions with underactuated objects~\cite{Li2026haic}. ResMimic refines a general motion tracker for loco-manipulation through residual learning~\cite{zhao2025resmimic}, while SceneBot conditions whole-body tracking on contact prompts for terrain and object interaction~\cite{scenebot2026}.

Complementary studies expand interaction data through human videos, retargeting, and motion generation~\cite{hdmi2025, sugar2026,wang2026humanx,yang2026omniretarget}. GRAIL generates interaction data and trains visuomotor policies from 3D assets and video priors~\cite{grail2026}. Other systems address force adaptation, sequential contacts, and extended task execution~\cite{zhang2025falcon,zhang2024wococo,cao2026hiwet,lin2026lessmimic,yu2026omnicontact}. In contrast to these approaches, ReGATE targets reactive HOI from onboard observations, with user commands that can modify locomotion and contact timing during execution.

\begin{figure*}[!t]
    \centering
    \includegraphics[
        width=\textwidth,
        height=\textheight,
        keepaspectratio
    ]{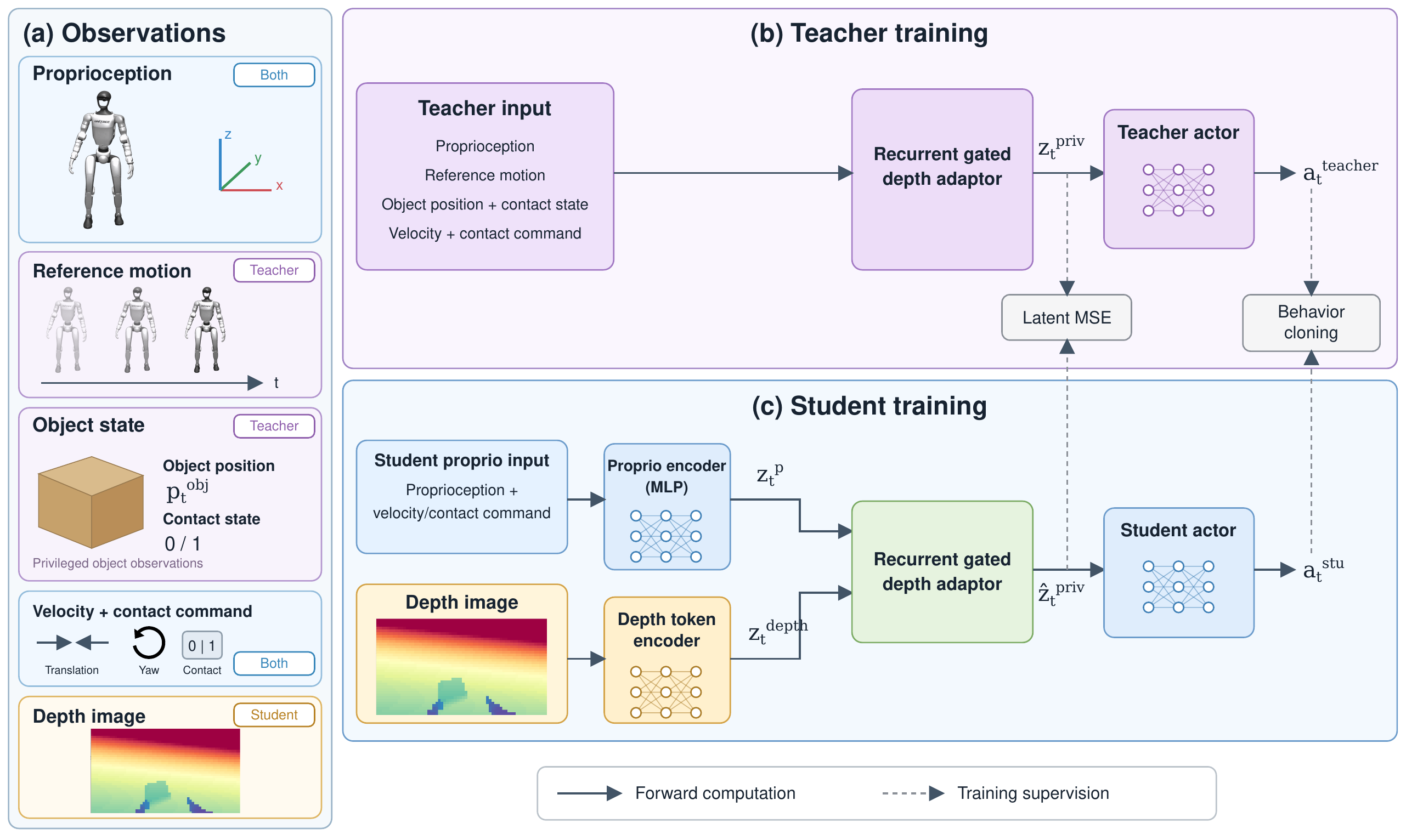}
    \caption{Overview of ReGATE.
    (a) Observations available to the teacher and student.
    (b) The privileged teacher produces the target latent
    $z_t^{\mathrm{priv}}$ and teacher actions.
    (c) The student combines proprioception, commands, and recurrent depth
    tokens to estimate $\hat z_t$ and produce joint targets. Dashed arrows denote training-only latent and action supervision.}
    \label{fig:regate_overview}
\end{figure*}

\subsection{Perceptive Humanoid Control}

Deployable HOI controllers must infer task-relevant object information from onboard sensors rather than simulator states. Modular approaches separate state estimation from control. PhysHSI employs a coarse-to-fine LiDAR--camera localization pipeline~\cite{wang2025physhsi}, whereas Pro-HOI~\cite{prohoi2026} combines online object detection with a digital twin for persistent estimation and uses FoundationPose~\cite{wen2024foundationpose} for object estimation. Such systems provide explicit geometric estimates, but their controllers depend on the accuracy and availability of an external perception pipeline.

Visuomotor approaches learn perceptual representations jointly with whole-body control. VisualMimic maps egocentric observations to commands for a task-agnostic motion tracker~\cite{visualmimic2025}, while GRAIL learns egocentric interaction policies from generated visual data~\cite{grail2026}. VAIC distills privileged interaction information into a depth-conditioned student driven by proprioceptive history and a decoupled velocity-and-contact command~\cite{vaic2026}, but still rely on object points template and reference motion command at implementation and deployment. ULTRA considers heterogeneous observation and goal modalities~\cite{ultra2026}, and RHINO uses human observations and commands to produce reactive humanoid--human--object interaction~\cite{chen2025rhino}. 

In ReGATE, we adopt the teacher--student backbone and command interface of VAIC, but construct its interaction representation within a recurrent adapter. The adapter predicts object state, posed geometry, and physical contact without requiring an external pose estimator or object template as a deployment input, providing a truly end-to-end solution.

\subsection{Exteroceptive--Proprioceptive Fusion}

Exteroception provides spatial information about the environment, whereas proprioception remains available when external observations are degraded or missing. Miki \emph{et al.} integrate proprioceptive and exteroceptive inputs through a recurrent belief state for perceptive locomotion~\cite{Miki2022}. AME and AME-2 apply proprioception-conditioned attention to robot-centric elevation maps to identify terrain regions relevant to locomotion~\cite{srAME, zhang2026ame2}. AMCO adapts multimodal coupling using semantic segmented RGB images~\cite{elnoor2024amco}, while CReF combines cross-attention, gated residual fusion, and recurrence for depth-conditioned humanoid locomotion~\cite{hao2026crefcrossmodalrecurrentfusion}. These methods demonstrate the value of adaptive perception fusion, but primarily address terrain-conditioned locomotion rather than contact-rich humanoid--object interaction.

Multimodal policies may also favor compact proprioceptive inputs over perceptual features required for target-dependent behavior~\cite{whenvisionfail}. ReGATE first grounds its depth representation through auxiliary geometry and contact supervision and then introduces visual degradation through curriculum training. Head-wise and global gates regulate how recurrent depth evidence updates an always-active proprioceptive representation throughout the interaction.

\section{Method}
\label{sec:method}

\subsection{Problem Formulation}
\label{sec:problem_formulation}

We consider whole-body humanoid--object interaction under onboard sensing. At time $t$, the deployable policy receives a proprioceptive observation $o_t^p$, an egocentric depth image $D_t\in\mathbb{R}^{1\times H\times W}$, and a causal command $u_t=[v_t^x,v_t^y,\omega_t^z,c_t]$. Here, $(v_t^x,v_t^y)$ and $\omega_t^z$ specify the desired planar velocity and yaw rate, while $c_t\in\{0,1\}$ indicates whether object contact is desired.  The policy
outputs joint-position targets $a_t\in\mathbb{R}^{23}$, which are tracked by a low-level proportional--derivative controller.

The object state is not directly observable by the deployed policy, and no external object-pose estimator is used. ReGATE instead maintains a recurrent interaction representation $\hat z_t$ from the history of proprioception, depth, and commands:
\begin{align}
    \hat z_t
    &=\mathcal R_{\theta}
      \!\left(o_{t-H+1:t}^{p},u_{t-H+1:t},D_{t-H+1:t}\right),
      \nonumber\\
    a_t
    &\sim\pi_S
      \!\left(\cdot\mid o_t^p,u_t,\hat z_t\right),
    \label{eq:deployable_policy}
\end{align}

where $\mathcal R_\theta$ denotes the recurrent gated fusion adapter and $\pi_S$ is the deployable student actor. 

As shown in Fig.~\ref{fig:regate_overview}, ReGATE employs the two-stage teacher--student framework. During training, a privileged teacher provides the latent target $z_t^{\mathrm{priv}}$ and action supervision. The student estimates this representation from deployable observations by combining recurrent depth features with an always-active proprioceptive pathway. The recurrent gated fusion adapter is detailed in Sec.~\ref{sec:recurrent_gate}, followed by the two-stage optimization procedure in Sec.~\ref{sec:teacher_student}.

\subsection{Recurrent Gated Depth Attention Adaptation Module}
\label{sec:recurrent_gate}
Depth is informative when the robot approaches an object and locates it, but its reliability varies after contact. Proprioception provides a complementary signal once the robot and object become dynamically coupled. ReGATE treats depth as a state-dependent correction to an always-available proprioceptive representation rather than as an unconditional policy input.

\subsubsection{Recurrent Spatial Depth Tokens}

We augment the depth image $D_t$ with normalized image-coordinate maps
$U,V\in[-1,1]^{H\times W}$ and encode the resulting three-channel tensor with a CNN. A spatial ConvGRU~\cite{ballas2016convgru} integrates consecutive feature maps, preserving both image location and temporal context. The recurrent feature map is projected into spatial depth tokens $\{z_{t,i}^{\mathrm{depth}}\}_{i=1}^{N}$ for subsequent attention.

\subsubsection{Proprioception-Conditioned Cross-Modal Attention}

An MLP encodes the proprioceptive observation $o_t^p$ and causal command $u_t$
into a control-context feature $z_t^p$. A 8-head cross-attention module uses $z_t^p$ as its query and the recurrent depth tokens as keys and values, producing the attention map $\mathcal A_t$ and head-wise visual evidence. The command-conditioned query makes visual retrieval dependent on the current robot state and interaction phase, while $z_t^p$ forms the residual path for the subsequent gated fusion.

\subsubsection{Hierarchical Gated Fusion}

\begin{figure}[t]
    \centering
    \includegraphics[
        width=0.99\columnwidth,
        clip
    ]{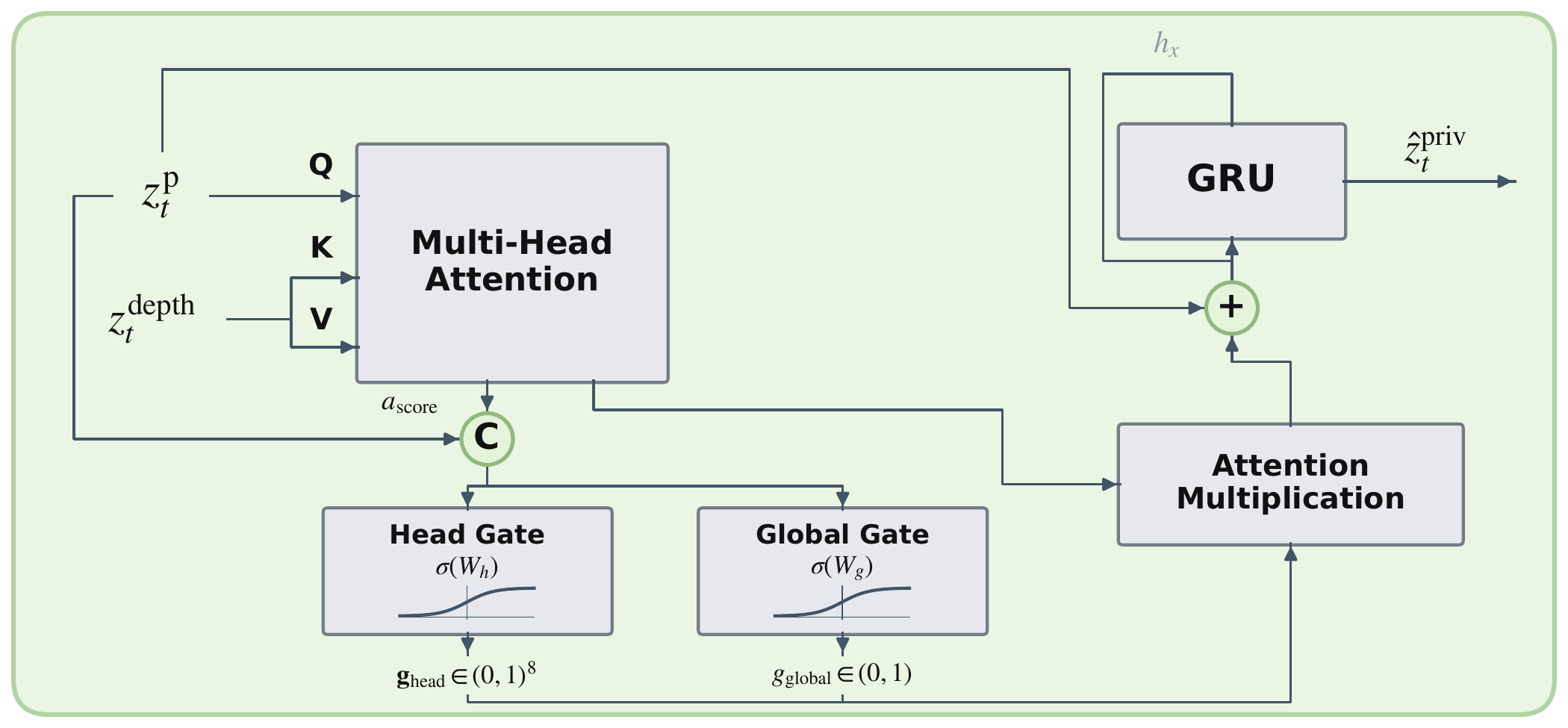}
    \caption{Architecture of the recurrent gated depth attention adapter. The proprioceptive feature $\mathbf{z}_t^{\mathrm{p}}$ forms the attention query, while the depth tokens $\mathbf{z}_t^{\mathrm{depth}}$ provide the keys and values.}
    \label{fig:gated_depth_attention}
\end{figure}

Depth observations may be occluded, missing, delayed, or repeated. Injecting every attended feature into the interaction belief can degrade a controller that already has a useful proprioceptive estimate. To limit harmful visual updates, ReGATE applies hierarchical gating to the attended depth evidence, following prior gated-fusion approaches~\cite{qiu2025gated,Miki2022}. Let $\boldsymbol{\mathcal{A}}_t^M=[\mathcal{A}_t^1;\ldots;\mathcal{A}_t^M]$ denote the evidence produced by attention head. A vector of head gates $g_t^{\mathrm{head}}\in(0,1)^M$ selects components of the attended evidence, while a scalar gate $g_t^{\mathrm{global}}\in(0,1)$ controls the overall depth correction. The fused representation is
\begin{align}
    z_t^{\mathrm{fuse}}
    &=\operatorname{LN}(z_t^{\mathrm{p}}+g_t^{\mathrm{global}}(g_t^{\mathrm{head}}\cdot \boldsymbol{\mathcal{A}}_t^M)).
    \label{eq:gated_fusion}
\end{align}
where the head-wise gates modulate individual visual subspaces and the global gate scales the complete visual correction. When $g_t^{\mathrm{global}}=0$, the representation reduces to the proprioceptive residual path. These gates are learned task-dependent coefficients rather than calibrated estimates of sensor reliability. This module is illustrated in Fig.~\ref{fig:gated_depth_attention}

\subsubsection{Interaction-State Prediction}

ReGATE estimates interaction-relevant object information without an external object-pose estimator. From $z_t^{\mathrm{fuse}}$, auxiliary heads predict the object state $\hat s_t^{\mathrm{obj}}$, posed object point cloud $\hat{\mathcal P}_t$, and physical end-effector contact $\hat c_t^{\mathrm{con}}$. The predicted geometry and contact, rather than $z_t^{\mathrm{fuse}}$ itself, condition the adaptation GRU:
\begin{equation}
    (\hat z_t,h_t^a)
    =
    \operatorname{GRU}_a\!\left(
    [o_t^p;u_t;\hat{\mathcal P}_t;\hat c_t^{\mathrm{con}}],
    h_{t-1}^a
    \right),
    \label{eq:adapt_gru}
\end{equation}
where $\hat z_t$ is the interaction latent supplied to the student actor. This design routes visual information through explicit geometry and contact estimates while recurrently aggregating the coupled robot--object state.

A depth-only head additionally predicts the image-space object location,
$\hat x_t^{2\mathrm D}=H_{2\mathrm D}(z_t^{\mathrm{depth}})$, from the depth tokens $z_t^{\mathrm{depth}}$. Because this head does not receive proprioception, its supervision directly grounds the visual representation during training.

\subsubsection{Depth Curriculum}
\label{sec:depth_curriculum}

Jointly training the gates and a randomly initialized depth encoder can cause premature gate closure as the controller can reduce its prediction error by suppressing visual features before they become informative. We address this problem with a visual-grounding phase. We begin with clean depth and force both gate levels open, allowing the adaptation losses to establish task-relevant visual features.

After this warm-up, the gates transition gradually to their learned values while structured occlusion and full-frame blackout are introduced. Training under both clean and corrupted observations encourages the adapter to exploit available depth without depending on it when visual evidence is missing.

\subsection{Two-Stage Teacher--Student Training}
\label{sec:teacher_student}

ReGATE follows the asymmetric teacher--student paradigm used in prior humanoid interaction controllers~\cite{vaic2026,Li2026haic}. The two stages separate acquisition of the interaction skill from learning its onboard perceptive realization.

\subsubsection{Stage I: Privileged Teacher and Student Warm-up}

Similar to ~\cite{vaic2026, Li2026haic}, we first train a privileged teacher $\pi_T$ with PPO using reference motions $s_t^{ref}$, ground-truth object states $s_t^{obj}$, contact information $c_t^{con}$. In teacher training, we initialize a privileged encoder using the same Recurrent Gated Depth Attention Adaptor as in Sec.~\ref{sec:recurrent_gate}, producing the latent target $z_t^{\mathrm{priv}}$. During teacher rollouts, the student adaptation module is warmed up with zero-valued depth features to reconstruct the privileged latent $\hat z_t$, object state $\hat s_t^{obj}$, posed object point cloud $\hat{\mathcal P}_t$, and physical contact state $\hat c_t^{\mathrm{con}}$ from proprioceptive information. This warm-up establishes a proprioceptive interaction prior before the depth encoder and learned gates are activated.

\subsubsection{Stage II: Visuomotor Student Training}

In Stage~II, we distill the student policy $\pi_S$ by receiving only onboard-compatible inputs of proprioception $o_t^p$, depth $D_t$, and the causal command $u_t$. Privileged quantities remain available in simulation only as supervision for the perception and adaptation modules. Their joint objective is
\begin{align}
    \mathcal L_{\mathrm{adapt}}
    ={}& \|\hat z_t-z_t^{\mathrm{priv}}\|_2^2
    +\lambda_o\|\hat s_t^{\mathrm{obj}}-s_t^{\mathrm{obj}}\|_2^2
    \nonumber\\
    &+\lambda_{\mathcal P}
    \|\hat{\mathcal P}_t-T(s_t^{\mathrm{obj}},\mathcal P)\|_2^2
    +\lambda_{2\mathrm D}
    \|\hat x_t^{2\mathrm D}-x_t^{2\mathrm{D, GT}}\|_2^2
    \nonumber\\
    &+\lambda_c\operatorname{BCE}
    \!\left(\hat c_t^{\mathrm{con}},c_t^{\mathrm{con, GT}}\right).
    \label{eq:adapt_loss}
\end{align}
The latent-alignment term distills the output of the frozen privileged encoder. The ground truth object-state target $s_t^{\mathrm{obj}}$ and the posed-cloud target are from the motion dataset. The image-space target $x_t^{2\mathrm D}$ directly supervises the depth tokens, while $c_t^{\mathrm{con}}$ records physical end-effector--object contact obtained from simulator and is distinct from the desired-contact command $c_t^{cmd}$ or the demonstration object-contact label $c_t^{\mathrm{obj}}$.

The student actor is additionally optimized to match the teacher action mean using $\mathcal L_{\mathrm{act}}=\|\mu_S-\mu_T\|_2^2$. Consequently, action distillation updates only the student actor, whereas the depth encoder, gated-fusion module, and adaptation network are optimized through $\mathcal L_{\mathrm{adapt}}$. The student actor is further refined by PPO during Stage~II.

\subsubsection{Phase-Aware Random State Initialization}

Uniform random state initialization (RSI) samples each interaction phase in proportion to its duration, underrepresenting short contact transitions. Phase-aware RSI instead assigns phase-specific sampling probabilities independent of phase length.

\begin{figure}[t]
    \centering
    \includegraphics[width=\columnwidth]{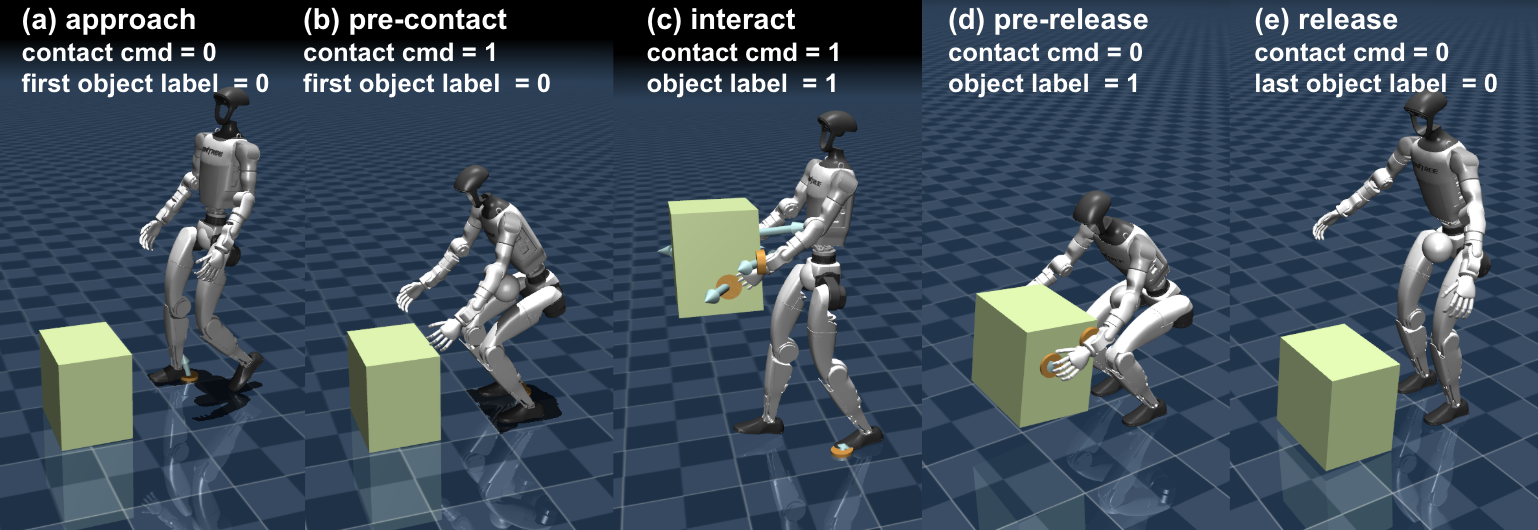}
    \caption{Phase-aware random state initialization. Demonstration frames are partitioned into five phases using the commanded contact state $c_t^{\mathrm{cmd}}$ and object-contact label $c_t^{\mathrm{obj}}$: (a) approach, (b) pre-contact, (c) interact, (d) pre-release, and (e) release. Approach and release are distinguished by their temporal positions before the first and after the final object contact.}
    \label{fig:phase_rsi}
\end{figure}

As shown in Fig.~\ref{fig:phase_rsi}, demonstrations are partitioned using the commanded contact state $c_t^{\mathrm{cmd}}$ and object-contact label $c_t^{\mathrm{obj}}$: \emph{approach} $(0,0)$ before the first contact, \emph{pre-contact} $(1,0)$, \emph{interact} $(1,1)$, \emph{pre-release} $(0,1)$, and \emph{release} $(0,0)$ after the final contact. Each reset first samples a phase from a task-specific distribution and then selects a frame uniformly within that phase, increasing coverage of transitions undersampled by uniform RSI.

\subsection{Perception--Action Rewards}
\label{sec:perception_action_rewards}

We retain the motion-tracking and locomotion-regularization rewards from prior work~\cite{hdmi2025,vaic2026,liao2025beyondmimic} and add lightweight, command-conditioned interaction rewards in Stage~II~\cite{grail2026,kong2026learningsoccerskillshumanoid}: (i) \emph{end-effector proximity} penalizes distance to the designated grasp points; (ii) \emph{object proximity} keeps the object within a reachable pelvis-centered region; (iii) \emph{facing alignment} encourages the robot to face the object; (iv) \emph{contact correctness} rewards commanded end effectors that establish physical contact; and (v) \emph{contact recovery} guides end effectors with missing contact back toward their grasp points. Together, these terms connect object-relative observations to the actions required for object interaction.

\section{Experiments}
\label{sec:experiments}

\subsection{Experiment Setup}
\label{sec:experiment_setup}

\subsubsection{Implementation Details}

We train ReGATE in Isaac Sim using the two-stage framework in Sec.~\ref{sec:teacher_student}. Both stages use PPO~\cite{ppo2017}, and the student adapter is optimized with Eq.~\eqref{eq:adapt_loss}. The teacher uses 4096 parallel environments, whereas ray-cast depth limits student training to 512. All experiments are trained on one NVIDIA GeForce RTX~5090 GPU.

The policy runs at $50\,\mathrm{Hz}$ and receives $64\times36$ depth images at $15\,\mathrm{Hz}$, with each frame held until the next visual update. Training randomizes depth noise, bias, latency, occlusion, blackout, and camera parameters to reduce the simulation-to-real sensing gap.

We augment the VAIC interaction motions~\cite{vaic2026} with an object-contact label $c_t^{\mathrm{obj}}$ and a desired-contact command $c_t^{\mathrm{cmd}}$. The command is activated approximately 60--70 frames before demonstrated contact, giving the policy time to approach and align its end effectors.

\subsubsection{Real-World Deployment Setup}
\label{sec:hardware_setup}

We deploy the student policy on a Unitree G1 humanoid with 29 actuated degrees of freedom. The policy predicts position targets for 23 joints while the six wrist joints are excluded from the policy action and held at fixed zero-position. 

Egocentric depth is captured by a head-mounted ZED~2i stereo camera at $30\,\mathrm{Hz}$. The deployment pipeline uses a field of view of approximately $101.4^{\circ}\times69^{\circ}$ and preprocesses each depth frame to the $64\times36$ input expected by the student. To match the training cadence, new depth observations are admitted at $15\,\mathrm{Hz}$ and held between updates, while proprioception and policy inference continue at $50\,\mathrm{Hz}$. The deployed policy receives no external motion-capture measurements, object-pose estimates, or simulator states during deployment.

\subsubsection{Task Description}
\label{sec:task_desc}

We evaluate ReGATE on box interaction in MuJoCo~\cite{todorov2012mujoco} and on a physical Unitree G1. The complete task includes approach, grasp, lift, transport, stationary holding, release, and return to standing. Object positions are varied laterally over $[-0.5,0.5]\,\mathrm{m}$ to evaluate behavior beyond reference-motion replay. 

Online commands specify contact transitions and locomotion during transport. We additionally test sensing robustness using phase-specific depth blackout.

The experiments address three questions:
\textbf{Q1:} Can ReGATE execute stable transitions in response to locomotion and contact commands?
\textbf{Q2:} Can ReGATE localize and acquire the object from onboard depth without an external object-pose estimator?
\textbf{Q3:} Can ReGATE complete interactions when depth observations are degraded or temporarily unavailable?

\subsubsection{Evaluation Metrics}
\label{sec:evaluation_metrics}

We evaluate our network in terms of task success rate where a trial is successful if the robot grasps the box after the contact command, transports it without falling or unintended release, releases it on command, and returns to a stable stance. For quantitative evaluation in simulator, we define success as completion of the 13-s episode with a pelvis height above $0.7$ m and an end-effector--box contact force of at least $1$ N.

We explicitly exclude from this work the evaluation of the approach to the object and the trigger decision for contact. Our end-to-end HOI policy focuses on the actual object interaction, while a vision-language model (VLM) or another higher-level controller would take care of the coarse approach and triggering without requiring precise motion control or a separate object pose estimator.

\subsection{Experimental Results}
\label{sec:experimental_results}

\subsubsection{\textbf{Q1}: Can ReGATE execute stable transitions in response to locomotion and contact commands}
\label{sec:result_q1}

\begin{figure}[t]
    \centering
    \includegraphics[width=\linewidth]
        {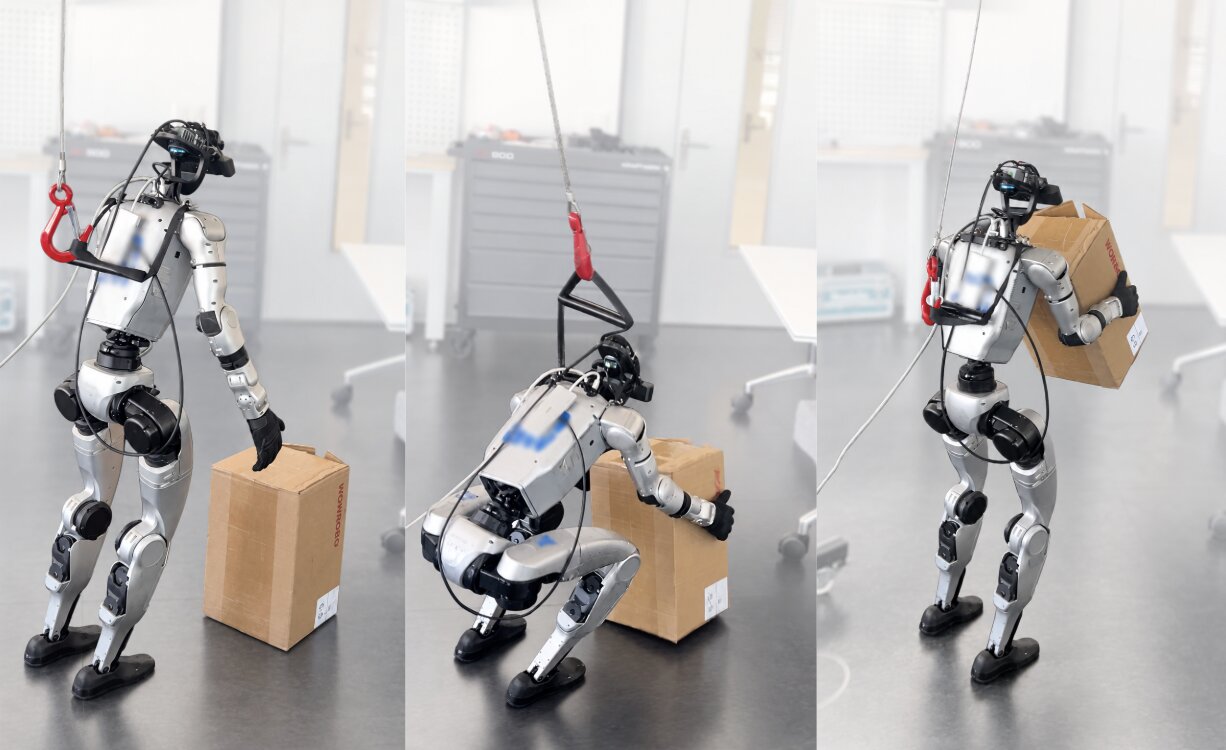}
    \caption{Side-offset box pickup with the real robot. The box is placed to the side of the robot when the robot receives the command to trigger contact with the box.}
    \label{fig:real_side_offset_pick}
\end{figure}

We evaluate whether ReGATE can execute a complete interaction sequence under online command changes. Each trial starts with the box placed in front of the robot. As shown in Fig.~\ref{fig:real_world_command} and Fig.~\ref{fig:real_side_offset_pick}, we activate sequential user commands and our policy can react accordingly.

ReGATE successfully completes all ten trials. During each trial, the robot transitions actively across each of the interaction stages while maintaining balance and object contact. It also returns to a stable standing posture after releasing the box. 

\subsubsection{\textbf{Q2}:Can ReGATE localize and acquire the object from onboard depth without an external object-pose estimator}

We evaluate whether ReGATE can locate and acquire objects using only proprioception and egocentric depth by placing boxes in different starting positions. Qualitative examples on the physical robot are shown in Fig.~\ref{fig:real_side_offset_pick}, where the policy grasps boxes placed at different lateral positions. These examples illustrate the complete onboard perception-to-action behavior without an external object-pose estimator.

For quantitative evaluation, we conduct two controlled studies in MuJoCo using the same policy as in deployment. The commanded velocity remains zero throughout both studies. In the static suite, the box is placed at 11 lateral offsets spanning $[-0.5,0.5]\,\mathrm{m}$, with 30 trials per position. We perturb the box's initial position and orientation by random offsets of up to $0.02\,\mathrm{m}$ and $0.05\,\mathrm{rad}$, respectively. In the moving suite, the box traverses the same lateral range at a sampled nominal speed between $0.1\,\mathrm{m/s}$ and $0.5\,\mathrm{m/s}$. Its lateral motion stops at the first physical end-effector contact, after which the robot must stabilize the freely moving robot--object system.

\begin{figure}[t]
    \centering
    \includegraphics[width=\linewidth]
        {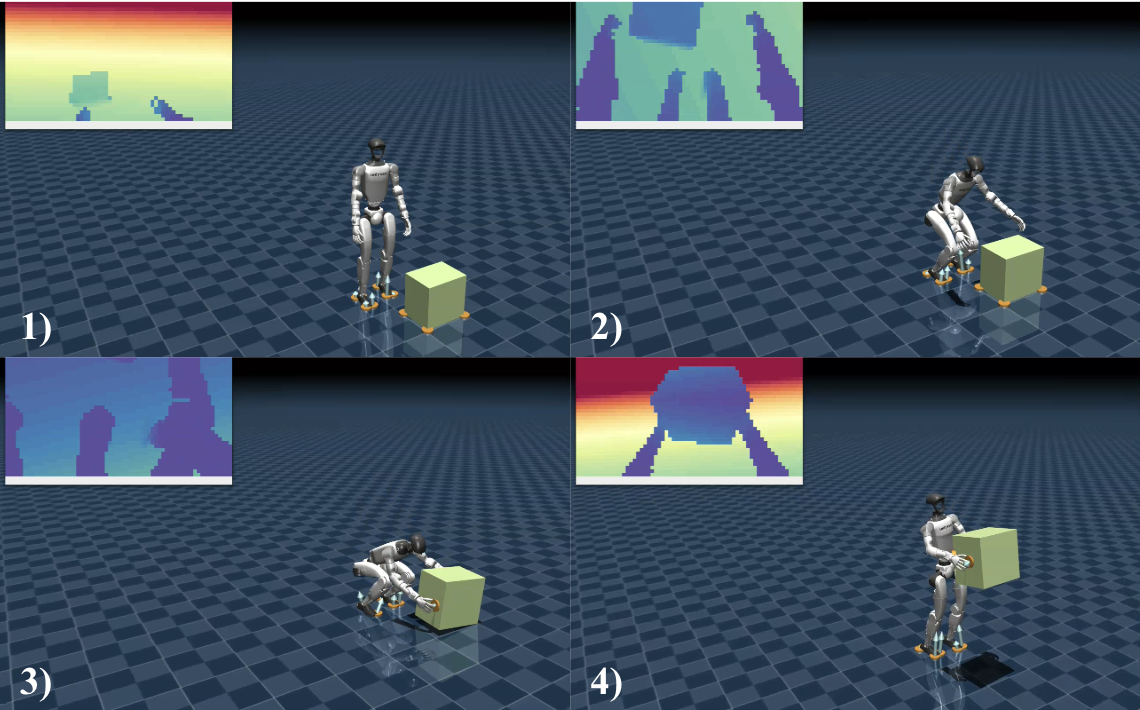}
    \caption{Side-offset box pickup in simulation using ReGATE. The box is placed $0.2\mathrm{m}$ left of the robot.
    (1) The robot starts from an upright stance.
    (2) Upon receiving the interaction command, it reaches toward
    the laterally offset box.
    (3) It establishes a two-handed grasp.
    (4) It lifts the box and returns to an upright stance for a
    standing hold. Insets show the corresponding egocentric depth
    observations.}
    \label{fig:side_offset_pick}
\end{figure}

ReGATE succeeds in 281 of 330 static trials, corresponding to an overall
success rate of 85.2\%. As shown in Table~\ref{tab:static_location}, the policy completes all 120 trials within the contiguous central region
$y\in[-0.2,0.1]\,\mathrm{m}$. Performance decreases near the lateral workspace limits, with the lowest success rates observed between $+0.2$ and
$+0.4\,\mathrm{m}$. Fig.~\ref{fig:side_offset_pick} illustrates a representative simulated pickup with a laterally offset box.

\begin{table}[t]
    \centering
    \caption{\small Static-object acquisition for varying lateral offsets.}
    \label{tab:static_location}
    \scriptsize
    \begin{tabular*}{\columnwidth}{
        @{\extracolsep{\fill}}lcccccc@{}
    }
        \toprule
        Box $y$ [m]
        & $-0.5$ & $-0.4$ & $-0.3$ & $-0.2$ & $-0.1$ & $0.0$ \\
        \midrule
        Success
        & 21/30 & 26/30 & 27/30 & 30/30 & 30/30 & 30/30 \\
        Rate [\%]
        & 70.0 & 86.7 & 90.0 & 100.0 & 100.0 & 100.0 \\
        \bottomrule
    \end{tabular*}

    \vspace{2pt}

    \begin{tabular*}{\columnwidth}{
        @{\extracolsep{\fill}}lcccccc@{}
    }
        \toprule
        Box $y$ [m]
        & $+0.1$ & $+0.2$ & $+0.3$ & $+0.4$ & $+0.5$ & Overall \\
        \midrule
        Success
        & 30/30 & 23/30 & 19/30 & 20/30 & 25/30 & 281/330 \\
        Rate [\%]
        & 100.0 & 76.7 & 63.3 & 66.7 & 83.3 & 85.2 \\
        \bottomrule
    \end{tabular*}
\end{table}

We compare ReGATE with VisualMimic~\cite{visualmimic2025} under the same static-object protocol. VisualMimic uses its official MuJoCo deployment configuration, with only the initial box position varied to match our evaluation. As reported in Table~\ref{tab:static_baseline}, ReGATE succeeds in 281 of 330 trials ($85.2\%$), compared with 160 of 330 ($48.5\%$) for VisualMimic, and achieves a higher success rate at every tested position. ReGATE adjusts its reaching motion to the various box locations and remains stable after lifting. In the observed VisualMimic failures, the end effectors either missed the box, preventing valid contact, or the robot lost stability while holding it after lifting.

\begin{table*}[t]
    \centering
    \setlength{\tabcolsep}{4pt}
    \scriptsize
    \caption{Comparison of static-object acquisition between ReGATE and
    VisualMimic in MuJoCo. \\Each lateral offset result is the average of 30 trials, for a total of 330 executions.}
    \label{tab:static_baseline}
    \begin{tabular}{@{}lcccccccccccc@{}}
        \toprule
        & \multicolumn{11}{c}{Success rate [\%] by box position $y$ [m]}
        & Success rate [\%] \\
        \cmidrule(lr){2-12}
        Method
        & $-0.5$ & $-0.4$ & $-0.3$ & $-0.2$ & $-0.1$
        & $0.0$ & $+0.1$ & $+0.2$ & $+0.3$ & $+0.4$ & $+0.5$
        & Overall \\
        \midrule
        VisualMimic
        & 23.3 & 33.3 & 43.3 & 46.7 & 43.3
        & 63.3 & 56.7 & 70.0  & 53.3 & 50.0 & 50.0
        & 48.5 \\
        ReGATE (ours)
        & \textbf{70.0} & \textbf{86.7} & \textbf{90.0}
        & \textbf{100.0} & \textbf{100.0} & \textbf{100.0}
        & \textbf{100.0} & \textbf{76.7} & \textbf{63.3}
        & \textbf{66.7} & \textbf{83.3}
        & \textbf{85.2} \\
        \bottomrule
    \end{tabular}
\end{table*}

We further evaluate moving-object acquisition on a laterally moving box. The box traverses the lateral workspace under four speed intervals,
and the contact command is activated when the box crosses one of 11 prescribed positions spanning $y\in[-0.5,0.5]$ m. We conduct 10 trials for every speed--position combination. The result is reported in Table~\ref{tab:moving_evaluation}. ReGATE completes 378 of the 440 trials, corresponding to an overall success
rate of $85.9\%$.  

Because object position was not randomized during training, the static-offset and moving-object evaluations constitute zero-shot tests at unseen locations. The results provide task-level evidence that ReGATE can use onboard depth to localize and acquire objects beyond the training configuration.

\begin{table*}[t]
    \centering
    \setlength{\tabcolsep}{3.5pt}
    \scriptsize
    \caption{Moving object acquisition in MuJoCo. The contact command is activated when the box crosses the prescribed lateral position. Each speed--position cell reports the success rate over 10 trials. }
    \label{tab:moving_evaluation}
    \begin{tabular}{c*{12}{c}}
        \toprule
        & \multicolumn{11}{c}{Success rate [\%] by commanded contact position $y$ [m]}
        & Success rate [\%] \\
        \cmidrule(lr){2-12}
        Speed [m/s]
        & $-0.5$ & $-0.4$ & $-0.3$ & $-0.2$ & $-0.1$
        & $0.0$ & $+0.1$ & $+0.2$ & $+0.3$ & $+0.4$ & $+0.5$
        & Overall \\
        \midrule
        $[0.1,0.2)$
        & 60.0 & 70.0 & 60.0 & 50.0 & 100.0
        & 100.0 & 90.0 & 100.0 & 100.0 & 50.0 & 30.0
        & 73.6 \\
        $[0.2,0.3)$
        & 90.0 & 100.0 & 70.0 & 50.0 & 90.0
        & 90.0 & 90.0 & 100.0 & 100.0 & 100.0 & 80.0
        & 87.3 \\
        $[0.3,0.4)$
        & 90.0 & 70.0 & 100.0 & 60.0 & 100.0
        & 90.0 & 100.0 & 100.0 & 100.0 & 100.0 & 100.0
        & 91.8 \\
        $[0.4,0.5)$
        & 90.0 & 80.0 & 100.0 & 60.0 & 90.0
        & 90.0 & 100.0 & 100.0 & 90.0 & 100.0 & 100.0
        & 90.9 \\
        \midrule
        Overall
        & 82.5 & 80.0 & 82.5 & 55.0 & 95.0
        & 92.5 & 95.0 & 100.0 & 97.5 & 87.5 & 77.5
        & 85.9 \\
        \bottomrule
    \end{tabular}
\end{table*}

\begin{figure}[t]
    \centering
    \includegraphics[width=\linewidth]
        {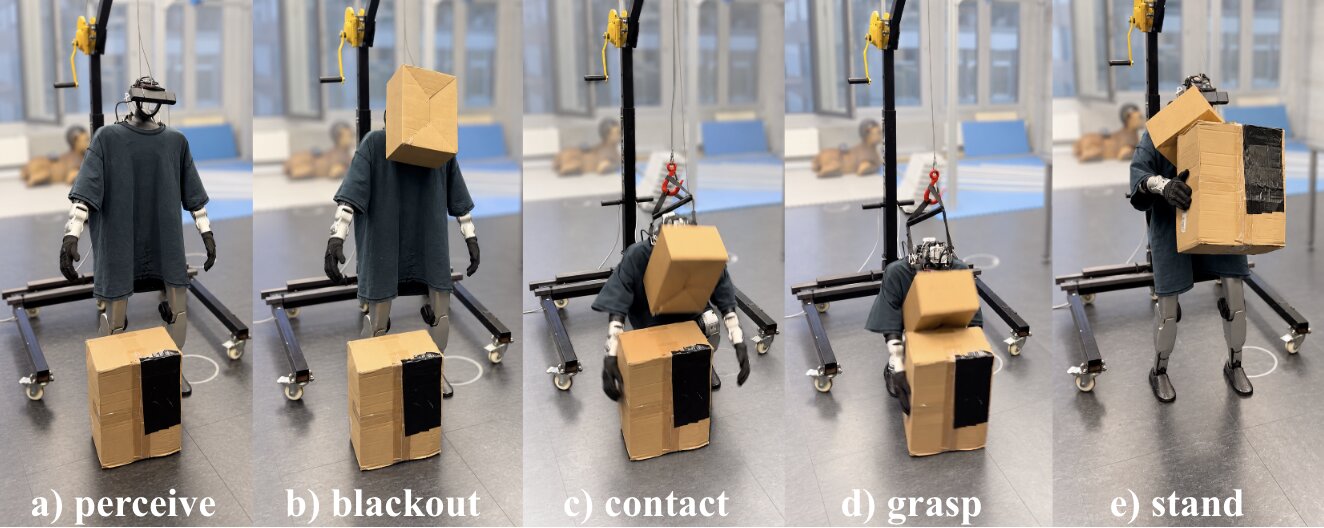}
    \caption{Sample successful demonstration where the real G1 robot picks up the box even when the vision input is completely blocked.}
    \label{fig:blackout_pick}
\end{figure}

\subsubsection{\textbf{Q3}: Can ReGATE complete interactions under degraded or unavailable depth}
\label{sec:results_degradation}

We introduce partial occlusion and full-frame blackout before contact, during transport, and before release, restoring clean depth without resetting the policy or its recurrent states. This protocol evaluates both the ability to maintain interaction during visual degradation and to resume visual updates after sensing recovers.

As demonstrated in Fig. \ref{fig:blackout_pick}, ReGATE maintains commanded behavior during temporary depth loss and resumes visually guided interaction after clean observations return. These results support the effectiveness of the recurrent gated fusion adapter, which maintains the interaction representation through its persistent proprioceptive pathway during visual degradation and reintroduces depth evidence through the learned gates when reliable observations return.

\section{Conclusion}
\label{sec:conclusion}

We presented ReGATE, a deployable whole-body controller for robust
humanoid--object interaction from onboard proprioception and egocentric
depth. ReGATE introduces a recurrent gated-fusion adapter that
regulates how depth updates a persistent proprioceptive representation.
Phase-aware random state initialization improves training coverage around
pre-contact and release transitions.

Experiments evaluate our framework under command-driven
interaction, onboard object localization and acquisition, and task
execution under degraded depth. The results demonstrate that the framework can infer object position from onboard sensing and reliably acquire objects without an external pose estimator. Under a matched static-object acquisition protocol, ReGATE achieves higher success than VisualMimic at every tested lateral offset. Moreover, recurrent gated fusion provides an effective mechanism for sustaining interaction behavior as the visual observations vary.


\section*{Acknowledgment}


This work was conducted in collaboration with Belimed AG as part of “Humanoid Robots for Cleaning and Sterilization of Medical Equipment” (project no. 134.504 IP-ICT), an innovation project supported by Innosuisse.

\bibliographystyle{IEEEtran}
\bibliography{references}

\end{document}